\documentclass{svjour3}                     
\smartqed  
\usepackage{graphicx}
\usepackage{amssymb}
\usepackage{natbib}
\newcommand{\eqref}[1]{(\ref{#1})}

\newcommand{\Str}{\mathrm{S}^+(3)}
\newcommand{\Su}{\mathbf{S}_1}
\newcommand{\Sd}{\mathbf{S}_2}
\newcommand{\Sm}{\mathbf{S}_{\mu}}
\newcommand{\Tt}{\mathrm{T}(3)}

\begin{document}

\title{An anisotropy preserving metric for DTI processing \thanks{This paper presents research results of the Belgian Network DYSCO (Dynamical Systems, Control, and Optimization), funded by the Interuniversity Attraction Poles Programme, initiated by the Belgian State, Science Policy Office. The scientific responsibility rests with its author(s).}
}


\author{Anne Collard         \and
        Silv\`{e}re Bonnabel \and
        Christophe Phillips \and
        Rodolphe Sepulchre 
}

\authorrunning{A. Collard \and S. Bonnabel \and C. Phillips \and R. Sepulchre} 

\institute{A. Collard, C. Phillips, R. Sepulchre \at
              Departement of Electrical Engineering and Computer Science,
    University of Li\`{e}ge, 4000 Li\`{e}ge, Belgium \\
              Tel.: +32-4366 3749\\
              \email{anne.collard@ulg.ac.be}           
           \and
           S. Bonnabel \at
              Robotics center, Math\'{e}matiques et Syst\`{e}mes,
    Mines ParisTech, Boulevard Saint-Michel 60, 75272 Paris, France 
    	\and
	C. Phillips \at
	Cyclotron Research Centre, 
    University of Li\`{e}ge, 4000 Li\`{e}ge, Belgium
}

\date{-}

\maketitle

\begin{abstract}
Statistical analysis of Diffusion Tensor Imaging (DTI) data requires a computational framework that is both numerically tractable (to account for the high dimensional nature of the data) and geometric (to account for the nonlinear nature of diffusion tensors). Building upon earlier studies that have shown that a Riemannian framework is appropriate to address these challenges, the present paper proposes a novel metric and an accompanying computational framework for DTI data processing. The proposed metric retains the geometry and the computational tractability of earlier methods grounded in the affine invariant metric. In addition, and in contrast to earlier methods, it provides an interpolation method which preserves anisotropy, a central information carried by diffusion tensor data. 

\end{abstract}

\section{Introduction}
\label{intro}
Diffusion-weighted imaging (DWI) allows non-invasive quantification of the self diffusion of water in vivo. In biological tissues, characterized by cell membranes and cytostructures, the movement of water is restricted because of these barriers. In tissues as white matter, which is highly directional, the resulting movement of water is therefore anisotropic. In this way, high diffusion anisotropy reflects the underlying highly directional arrangement of white matter fibre bundles. Diffusion measurements (which use the same tools as magnetic resonance imaging (MRI)) can characterize this anisotropy. The most common representation of this directional diffusion is through the use of diffusion tensors, a formalism introduced by Basser et al in 1994 \citep{Basser1994259}. Since then, other higher level representations have been introduced, as the Q-Ball Imaging \citep{MRM:MRM20279} and the Diffusion Kurtosis Imaging \citep{NBM:NBM1518}. In the context of Diffusion Tensor Imaging (DTI), each voxel of the image contains a diffusion tensor, which is derived from a set of DWI measured in different directions. A diffusion tensor is nothing else than a symmetric positive definite matrix whose general form is given by

\begin{equation}
\mathbf{D}= \left( \begin{array}{ccc} D_{xx} & D_{xy} & D_{xz}\\ D_{xy} & D_{yy}& D_{yz}\\ D_{xz} & D_{yz} & D_{zz}\end{array}\right) \end{equation} where $D_{xx} , D_{yy}, D_{zz}$ relate the diffusion flows to the concentration gradients in the $x,y$ and $z$ directions. The off-diagonal terms reflect the correlation between diffusion flows and concentration gradients in orthogonal directions. This diffusion tensor can be graphically represented as an ellipsoid. This ellipsoid takes the three eigenvectors of the matrix as principal axes (representing the three principal directions of diffusion). The length of the axes, related to the intensities of diffusion along them, is determined by the eigenvalues. Diffusion tensor images can thus be viewed as fields of ellipsoids. 

Classical image processing methods have been developed for scalar fields. As a result, early processing of DTI data first converted the tensor information into scalar data, for instance focusing on the scalar measure of fractional anisotropy (FA), see \emph{e.g.} \citep{Alexander2000233}. However, the tensor nature of DTI data soon motivated a generalization of signal processing methodological frameworks to tensor fields. Methods based on the Riemannian geometry of symmetric positive definite matrices have emerged in particular \citep{citeulike:451696,Fletcher07riemanniangeometry,Castano-Moraga2006,gur:fast,Batchelor:2005ys,Lenglet:2006uq,Lenglet2009S111} because the geometric framework provides a nonlinear generalization of calculus in linear spaces.\\
The present paper adopts the Riemannian framework to propose a novel metric and similarity measure between diffusion tensors. Our objective is to retain the geometry and computational efficiency of existing methods while addressing their main limitation, \emph{i.e.} the degradation of the anisotropy information through averaging and interpolation.\\
Our proposed similarity measure provides a concept of averaging and a concept of anisotropy that have the remarkable property to commute: the anisotropy of the average tensor is the average of the anisotropies. In that sense, our similarity measure is anisotropy-preserving. \\
To arrive at this desirable property, we need a metric that compares separately the orientation of the tensors and their spectrum. Earlier methods based on this spectral decomposition suffer a strong computational obstacle that we remove thanks to quaternions. The use of quaternions to compare orientations is analog to the use of Log-Euclidean computation to compare positive definite matrices. In that sense, our spectral quaternions framework for computation is analog to the Log-Euclidean framework for computation in the affine invariant geometry \citep{arsigny:328}, with similar computational gains. \\
The paper is organized as follows: Section 2 reviews the geometry of diffusion tensors and introduces the novel metric. Section 3 develops the computational framework, that is a similarity measure that captures the desired geometry while being computationally tractable. Section 4 introduces the desired concepts of mean and anisotropy, which are shown to commute. Section 5 illustrates the properties of the proposed framework. Section 6 contains concluding remarks.


\section{Riemannian metric for diffusion tensor processing}
\label{sec:1}
Even though the statistical analysis of diffusion tensor images still largely relies on scalar quantities such as the fractional anisotropy (FA) \citep{Alexander2000233}, the potential benefit of properly exploiting the tensor information of these data has been recognized early. The most common matrix representation of diffusion tensor is by elements of $\mathrm{S}^+(3)$, the set of $3 \times 3$ symmetric positive definite matrices.\\
Because this set is a nonlinear space, basic processing operation such as averaging and interpolation cannot be performed in the usual (euclidean) way. Defining and computing such quantities in nonlinear spaces equipped with a Riemannian geometry is a topic of active ongoing research, motivated by many applications in statistical signal processing \citep{Smith05_cov,citeulike:451696,Ando2004305,Petz:2005:MPN:1103420.1108724,springerlink:10.1007/s10851-010-0255-x,doi:10.1137/S0895479803436937,Burbea1982575,Skovgaard84}. Focusing on $\mathrm{S}^+(3)$, we briefly review the importance of the affine-invariant metric and we introduce a novel metric. 

\subsection{The affine invariant Riemannian geometry of positive definite matrices}
\label{sec:2}
A Riemannian framework for DTI processing was first introduced in \citep{citeulike:451696} and \citep{Fletcher07riemanniangeometry}. It is based on the parametrization of the space of symmetric definite positive matrices $\mathrm{S}^+(3)$ as a quotient space, \emph{i.e.}
\begin{equation} \mathrm{S}^+(3) \approx \mathrm{Gl}(3)/\mathrm{O}(3)\label{quo_geo} \end{equation}
where $\mathrm{Gl}(3)$ is the space of general linear matrices (representing all the possible affine transformations) while $\mathrm{O}(3)$ is the space of orthogonal matrices. In matrix term, any element $\mathbf{S} \in \mathrm{S}^+(3)$ can be represented by an invertible matrix $\mathbf{A} \in \mathrm{Gl(3)}$ through the factorization $\mathbf{S}= \mathbf{AA}^T$, but this representation is unique only up to multiplication by an orthogonal matrix $\mathbf{U} \in \mathrm{O(3)}$ because $\mathbf{AA}^T= (\mathbf{AU})(\mathbf{AU})^T$.\\
The quotient geometry \eqref{quo_geo} is reductive \citep{Smith05_cov}, which owes to the decomposition of an arbitrary $3\times 3$ matrix (a tangent vector to $\mathrm{Gl(3)}$) as the sum of a skew symmetric matrix (a tangent vector to $\mathrm{O}(3)$) and of a symmetric matrix (a tangent vector to $\mathrm{S}^+(3)$). As a consequence, the set $\mathrm{S}^+(3)$ can be equipped with the special metric
\begin{equation} g_{\mathbf{S}}(\xi, \eta) = \mathrm{tr}(\mathbf{S}^{-1/2} \xi \mathbf{S}^{-1} \eta \mathbf{S}^{-1/2}) \label{aff_metric} \end{equation} 
which is called affine-invariant because of the property
\[ g_{\mathbf{GSG}^T}(\mathbf{G}\xi \mathbf{G}^T,\mathbf{G}\eta \mathbf{G}^T) =g_{\mathbf{S}}(\xi, \eta), \qquad \forall \mathbf{G} \in \mathrm{Gl}(3)\, . \]
The resulting Riemannian distance has the explicit expression
\[ d(\mathbf{S}_1, \mathbf{S}_2) = || \log(\mathbf{S}_1^{-1/2} \mathbf{S}_2 \mathbf{S_1}^{-1/2}||\,, \]
and inherits the affine invariance property of the metric, that is
\begin{equation} d(\mathbf{GS}_1\mathbf{G}^T, \mathbf{G} \Sd \mathbf{G}^T) = d(\Su, \Sd)\, , \qquad \forall \mathbf{G} \in \mathrm{Gl}(3)\, . \label{aff_prop} \end{equation}
The metric invariance has two concrete consequences of importance for DTI processing. First, it is invariant to geometric transformation of the data such as rotation (\emph{i.e} independent of the (arbitrary) orientation of the laboratory axes) or scaling (independent of the choice of length units), a very desirable property that respects the physical nature of MRI data. Second, the distance to identity 
\begin{eqnarray}
 d(\mathbf{S}, \mathbf{I}) & = & || \log \mathbf{S} || \\
  &=& \sqrt{\sum_i \log ^2 \lambda_i }
  \label{dist_ide}
\end{eqnarray}
is indeed a measure of anisotropy, that is, does not depend of the orientation of the tensor $\mathbf{S}$ but only on its spectrum. The affine invariant framework has been used in several works in the context of DTI processing \citep{Fletcher2009S143,Castano-Moraga2006,gur:fast,Batchelor:2005ys,Lenglet:2006uq,Lenglet2009S111}.

\subsection{A novel Riemannian metric based on the spectral decomposition}
\begin{figure}[ht]
\begin{center}
\includegraphics[scale=0.35]{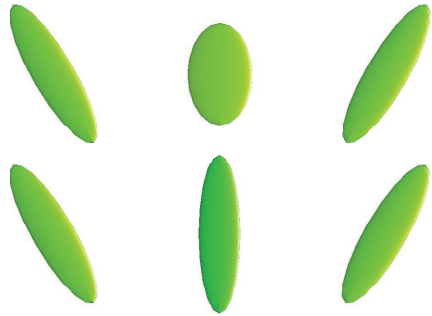}
\caption{Mean of two cigar-shaped diffusion tensors. \emph{Top:} Affine-Invariant mean \citep{citeulike:451696}, \emph{bottom:} mean computed with the method proposed in this paper, the spectral quaternions method. }
\label{fig1}
\end{center}
\end{figure}

A main motivation for the present paper is to address a limitation of the affine invariant metric graphically illustrated in Figure \ref{fig1}, top: the anisotropy of tensors tends to be smoothened and reduced through averaging. If anisotropy is a key information of tensor, this means a loss of information that can quickly accumulate through the many averaging operations involved in data processing. This limitation, intrinsically related to the affine invariance of the metric \eqref{aff_metric}, has motivated alternative similarity measures that compare orientation and spectral information separately \citep{Weldeselassie:2009fk,10.1109/CVPR.2001.990631,Ingalhalikar:2010vn}, starting from the spectral decomposition 

\begin{equation}
\mathbf{S}= \mathbf{U} \Lambda \mathbf{U}^T,
\label{dec_spe}
\end{equation}
where $\mathbf{U}$ is an orthogonal matrix containing the eigenvectors of the tensor and $\Lambda$ is a diagonal matrix whose entries are the eigenvalues of the tensor. Since tensors are symmetric positive definite matrices, $\Lambda \in \mathrm{D}^+(3)$, the space of diagonal matrix with positive elements, and the matrix $\mathbf{U}$ belongs to the orthogonal group $\mathrm{O}(3)$. In this work, we will order the eigenvalues of the tensors such that $\lambda_1\geq \lambda_2 \geq \lambda_3$ and we will impose $\det(\mathbf{U})=1$, i.e., $\mathbf{U}$ belongs to the space of special orthogonal matrices, $\mathrm{SO}(3)$. \\

Figure \ref{fig1}, bottom, shows an example of a mean computed using the method developed in this work, and compares it to the affine-invariant mean (top). Both methods preserve the determinants of tensors (linked to the volume of ellipsoids) but, in contrast to the affine-invariant mean, the proposed mean conserves the shape of the tensor. This difference is of importance in the context of Diffusion Tensor Imaging, as the anisotropy property is a key information for instance in tractography. The following subsection will introduce in more details the proposed metric. \\

A Riemannian metric based on the spectral decomposition \eqref{dec_spe} is obtained as follows. Our construction builds upon earlier work \citep{Bonnabel:2009uq} that aims at extending the affine invariant metric to 'flat' ellipsoids, that is to positive semidefinite matrices of given rank. \\

\begin{figure}[ht]
\begin{center}
\includegraphics[scale=0.5]{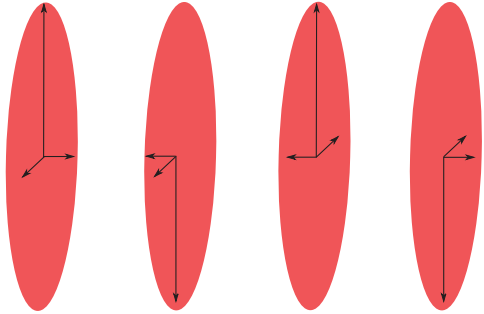}
\caption{Non-uniqueness of the spectral decomposition. Four different rotation matrices parametrize the same tensor. }
\label{overparam}
\end{center}
\end{figure}

Let ${\mathrm{S}_*}^+(3)$, be the subset of $\mathrm{S}^+(3)$ where all the eigenvalues are distinct, i.e., $0<\lambda_3< \lambda_2 < \lambda_1$. Each tensor $S$ of  ${\mathrm{S}_*}^+(3)$ can be identified to a diffusion ellipsoid whose equation is $\{x\in\mathbb R^3, x^TS^{-1}x= 1\}$. Any such ellipsoid is determined by three orthogonal axes, and three positive values corresponding to the length of the ellipsoid's axes. Note that the axes are not oriented. Thus, the following identification holds 
$${\mathrm{S}_*}^+(3)\approx \mathrm{T}(3) \times \mathrm{D}^+(3)$$where  $\mathrm{T}(3)$ is defined the following way: each element of $\mathrm{T}(3)$ is a set of three orthogonal \emph{directions}  in $\mathbb R^3$, i.e., $T(3)=\{(s_1v_1,s_2v_2,s_3v_3)\mid v_i.v_j=\delta_{ij}, (s_1,s_2,s_3)\in\mathbb R_*^3\}$. $\mathrm{D}^+(3)$ is the set of $3 \times 3$ diagonal matrices with positive entries. When two or more eigenvalues are equal, the tensor is described by an infinity of elements of $\mathrm{T}(3)$. This case almost never occurs in practice and is thus excluded.  The  set $\mathrm{T}(3)$ can be advantageously parameterized by matrices of $\mathrm{SO}(3)$ based on the identification $\mathrm{T}(3)\simeq \mathrm{SO}(3)/\mathrm{G}$ where $\mathrm{G}$ is the discrete group acting of $\mathrm{SO}(3)$ by multiplying two arbitrary columns jointly by $-1$ (i.e. $\mathrm{G}$ is the group of the three rotations of angle $\pi$ around the axes of the reference frame, plus the identity rotation). More prosaically, this  identification expresses the fact that  the orientation information contained in each tensor of ${\mathrm{S}_*}^+(3)$ can be described by 4 distinct rotation matrices, see Figure \ref{overparam}.        As a result this set can be viewed as a product ${\mathrm{S}_*}^+(3)\approx (\mathrm{SO}(3)/\mathrm{G}) \times \mathrm{D}^+(3)$. The space $\mathrm{SO}(3)/\mathrm{G}$ is locally isomorphic to $\mathrm{SO}(3)$ (we say $\mathrm{SO}(3)$ is a quadruple cover of this set in the topological sense) and inherits the usual metric  $g_{\mathrm{SO}(3)}$ of this latter space.  We thus propose to define a Riemannian metric on ${\mathrm{S}_*}^+(3)$ expressed on the tangent space at $S=U\Lambda U^T$ as follows 
\[ g_{\mathrm{S}^+(3)}^S= k(\Lambda) g_{\mathrm{SO}(3)}^U+ g_{\mathrm{D}^+(3)}^\Lambda\]
where $g_{\mathrm{SO}(3)}$ is the metric of the space $\mathrm{SO}(3)$ and $g_{\mathrm{D}^+(3)}$ the metric of $\mathrm{D}^+(3)$ to be defined later, and where  $k$ is a weighting factor. \\
The choice of the weighting factor $k$ is motivated by the following fact:  the more isotropic a tensor, the larger the uncertainty about its orientation \citep{Parker:2003fk}. On the contrary, if two very anisotropic tensors are compared, the information contained in their orientation is very precise and the discrepency between orientations must be emphasized. We thus choose the factor $k$ as  a function of the degree of anisotropy of the tensor. We let $k$ be equal to zero when all the eigenvalues are equal, and tend to 1 as the anisotropy becomes large. 

\section{Computationally tractable similarity measures}

Similarity measures for DTI should not only be geometrically meaningful. They must also be computationally tractable to scale with the combinatorial number of averaging operations involved in the statistical processing of 3D brain images.\\

Computational tractability motivated a simplification of the affine invariant metric known as the Log-Euclidean metric \citep{arsigny:328}. Due to its numerical efficiency, the Log-Euclidean metric has been used extensively in recent studies \citep{Goodlett2009S133,Chiang:2008fk,Ingalhalikar:2010vn,Castro:2007uq,Weldeselassie:2007ys,Arsigny:2006fk,Fillard:2007vn,Rodrigues:2009kx,Yeo:2009ly,Yeo:2008zr,Lepore:2006ve,Awate:2007qf}. The geometry of the Log-Euclidean metric simply exploits the property that the nonlinear space $\mathrm{S}^+(3)$ is mapped to the linear space $\mathbb{R}^{3 \times 3}$ by the (matrix) log mapping $\mathbf{S} \rightarrow \log \mathbf{S}$. 
The Log-Euclidean distance is thus 

\begin{equation}
d_{\mathrm{LE}}(\Su,\Sd) = || \log(\Su)-\log(\Sd)||_{\mathrm{F}}
\label{LEdist}
\end{equation}
where $||A||_\mathrm{F}$ is the Frobenius norm of matrices. This distance looses the affine invariance property \eqref{aff_prop} but retains the two concrete properties singled out in the previous section: \emph{(i)} invariance to rotations and scaling and \emph{(ii)} spectral nature of the distance to identity \eqref{dist_ide}. It suffers the same averaging limitation as the affine-invariant metric illustrated in Figure \ref{fig1}.

In an analog way, we will exploit the euclidean embedding of rotation matrices in the linear space of quaternions to arrive at a computationally tractable similarity measure. 

The proposed Riemannian metric provides an infinitesimal notion of length on the set ${\mathrm{S}_*}^+(3)$, which allows to measure the length of any curve in ${\mathrm{S}_*}^+(3)$. The geodesic distance between two tensors is defined as the infimum of the length of  curves joining those two tensors. Instead of calculating it, which can prove quite complicated and computationally expensive, we propose to directly define a similarity measure between tensors that approximates the geodesic distance, which is given in explicit form by

\begin{equation}
d^2_{\mathrm{SD}}(\Su,\Sd)= k(\Lambda_1, \Lambda_2)\, d^2_{\mathrm{SO}(3)}(\mathbf{U}_1,\mathbf{U}_2)+ d^2_{\mathrm{D}^+(3)}(\Lambda_1, \Lambda_2)
\label{sim_SD}
\end{equation}
where $d_{\mathrm{SO}(3)}(\mathbf{U}_1,\mathbf{U}_2)$ is a certain distance between two rotation matrices in the space $\mathrm{SO}(3)$, $\mathbf{U}_1$ and $\mathbf{U}_2$ are two well chosen representatives, while $ d_{\mathrm{D}^+(3)}(\Lambda_1, \Lambda_2)$ is a certain distance between two positive diagonal matrices, both supposed to approximate the geodesic distances in the respective spaces $\mathrm{SO}(3)$ and $\mathrm{D}^+(3)$. The function $k$ is supposed to account for the anisotropies of $\Su$ and $\Sd$.

\subsubsection*{Comparing intensities of diffusion}

As explained above, the similarity measure should be invariant under scalings and rotations. This implies that the distance between eigenvalues should be invariant to scalings (this distance is not affected by rotation). This is indeed the case as soon as  their distance is of the form
\begin{equation}
d_{\mathrm{D}^+(3)}(\Lambda_1,\Lambda_2)= \sqrt{\sum_i \log^2\left( \frac{\lambda_{1,i}}{\lambda_{2,i}} \right)}
\label{dist_D3}
\end{equation} 
where $\lambda_{1,i}$ and $\lambda_{2,i}$ are the eigenvalues of $\Su$ and $\Sd$, respectively.

With the choice \eqref{dist_D3}, the distance \eqref{sim_SD} to identity coincides with \eqref{dist_ide}, provided that the weighting factor $k$ tends to zero whenever $\Su$ or $\Sd$ becomes isotropic.


\subsubsection*{Comparing orientation of diffusion}

A longstanding computational problem of the spectral approach for processing of tensor images is the overparameterization of $\mathrm{T}(3)$ by orthogonal matrices  (see Figure \ref{overparam}), as we have seen that $\mathrm{T}(3)$ admits a quadruple cover by $\Str$.  In preceding works using the spectral decomposition \citep{10.1109/CVPR.2001.990631,chefdhotel04}, methods of \emph{realignment} of rotation matrices have been proposed to overcome this  difficulty. Mathematically, if we denote by $\mathcal{U}_i$ the set of four rotation matrices representing the orientation of the tensor $\mathbf{S}_i$, these methods search for the pair of rotation matrices $\mathbf{U}_1$ and $\mathbf{U}_2$ which minimizes the distance between $\mathcal{U}_1$ and $\mathcal{U}_2$

\begin{equation}
d(\mathcal{U}_1, \mathcal{U}_2)= \min_{\mathbf{U}_2 \in \mathcal{U}_2} d_{\mathrm{SO}(3)}(\mathbf{U}_1^r, \mathbf{U}_2) 
\label{dist_rot}
\end{equation}
where $\mathbf{U}_1^r$ is the chosen reference in $\mathcal{U}_1$ and $d_{\mathrm{SO}(3)}(\mathbf{U}_1^r, \mathbf{U}_2)$ is the geodesic distance in the $\mathrm{SO}(3)$ manifold
\begin{equation}
d_{\mathrm{SO}(3)}(\mathbf{U}_1, \mathbf{U}_2)= ||\log(\mathbf{U}_1^T \mathbf{U}_2)||.
 \label{distso} 
 \end{equation}

This step can thus be expressed in our framework as the search for the geodesic distance in $\Tt$ between two orientations in the sense of the natural metric of $\Tt$, i.e., the metric inherited from its covering group $\mathrm{SO}(3)$. 
However, since this realignment is local and has to be performed for each tensor of any image, this procedure dramatically increases the computational cost of processing algorithms. 

It is thus the parametrization of diffusion tensors by rotation matrices that make the framework computationally expensive. But the orthogonal group is mapped to a linear space through the log operation, exactly as the space $\Str$. This is the essence of the classical quaternion-based computational framework of rotations.

\subsubsection*{Euclidean embedding of orientations}
The use of quaternions is very popular in robotics to efficiently compute over rotation matrices. The group of quaternions of norm 1, usually denoted $\mathbb H_1$ provides an universal cover of $\Str$ and thus an universal cover of $\Tt$.  
A unit quaternion is generally denoted by $\mathbf{q}= (a, \mathbf{V})$ where $a$ is associated to the angle of rotation by $\theta= 2 \arccos(a)$ and $\mathbf{V}$ is associated to the axis $\mathbf{w}$ of rotation through $\mathbf{w}= \mathbf{V}/ \sin(\theta/2)$. From $\mathbf{q}$, the associated rotation matrix $\mathbf{R}$ is given by
\begin{equation}
\mathbf{R}= \exp \left( \begin{array}{ccc} 0 & -w_3 \theta & w_2 \theta \\ w_3 \theta & 0 & -w_1 \theta \\ -w_2 \theta & w_1 \theta & 0 \end{array}\right). 
\label{quat2mat}
\end{equation}
The construction of $\mathbf{q}$ from $\mathbf{R}$ is given by \begin{equation}\theta= \arccos((\mathrm{trace}(\mathbf{R})-1)/2) \end{equation}
\begin{equation}
\mathbf{w}= \frac{1}{2\sin \theta} \left( \begin{array}{c} R_{3,2}-R_{2,3} \\ R_{1,3}- R_{3,1}\\ R_{2,1}-R_{1,2} \end{array}\right). 
\label{mat2quat}
\end{equation}
Finally, we have $a= \cos(\theta/2)$, $\mathbf{V}= \sin(\theta/2) \mathbf{w}$. Note that the opposite quaternion given by $(-a, -\mathbf{V})$ represents the same rotation matrix. Using this representation, rotations can be manipulated as Euclidean vectors, which decreases the computational cost. For example, we can use a simple Euclidean norm to compute the distance between two unit quaternions 

\begin{equation}
d(\mathbf{q}_1, \mathbf{q}_2) = ||\mathbf{q}_1- \mathbf{q}_2||\
\label{dist_quat}
\end{equation}

This is the so-called chordal distance on the sphere of quaternions, which is a good approximation of the geodesic distance of  $\mathrm{SO}(3)$ for close enough quaternions in $\mathbb H_1$. This latter group being an octuple cover of $\Tt$, processing tensor orientations with quaternions could seem more involved, but proves to be much more economical, computationally speaking. Indeed,  let $\mathcal{Q}_i$ be the set of eight quaternions representing the orientation of the tensor $\mathbf{S}_i$. We propose the following distance on $\Tt$
\begin{equation}
d(\mathcal{Q}_1, \mathcal{Q}_2)= \min_{\mathbf{q}_2 \in \mathcal{Q}_2} || \mathbf{q}^r_1 - \mathbf{q}_2||
\label{dist_Quat}
\end{equation}
where $\mathbf{q}^r_1$ is one element of $\mathcal{Q}_1$ chosen as a reference. This element can be chosen arbitrarily, and the distance is a good approximation of the geodesic distance on $\mathrm{T}(3)$, inherited by the geodesic distance on $\mathrm{SO}(3)$. Since $\mathbf{q}_1$ and $\mathbf{q}_2$ are of unit norm, this distance can be advantageously expressed as
\begin{equation}
|| \mathbf{q}_1-\mathbf{q}_2||^2= 2- 2\, \mathbf{q}_1. \mathbf{q}_2 
\label{dist_quat2}
\end{equation}
Using Eq.\eqref{dist_quat2}, the distance between $\mathcal{U}_1$ and $\mathcal{U}_2$ can be computed through
\begin{eqnarray}
\mathbf{q}_2^a &= &\max_{\mathbf{q}_2 \in \mathcal{Q}_2} \mathbf{q}_1^r.\mathbf{q}_2 \label{r_max}\\
d(\mathcal{Q}_1, \mathcal{Q}_2) &=& || \mathbf{q}_1^r-\mathbf{q}_2^a|| 
\label{dist_max}
\end{eqnarray}
where $\mathbf{q}_2^a$ is called the \emph{realigned} quaternion. 

Compared to \eqref{dist_rot}, the computation of \eqref{r_max} and \eqref{dist_max} is very fast. Indeed, the eight scalar products $\mathbf{q}_1^r.\mathbf{q}_2$ can be computed through a single matrix product between the $1 \times 4$ vector representing $(\mathbf{q}_1^r)^T$ and the $4 \times 8$ matrix formed by the eight quaternions $\mathbf{q}_2$. In contrast, computing the distance  \eqref{dist_rot} requires four logarithms of product of $3 \times 3$ matrices, which is expensive. The selection of the parametrization of rotations as quaternions thus enables the framework to be computationally tractable. Moreover, the distance corresponds to a natural distance on $\Tt$, and is thus invariant to rotations.\\

The similarity measure \eqref{sim_SD} between two tensors $\Su$ and $\Sd$ becomes
\begin{equation}
d^2_{\mathrm{SD}}(\Su,\Sd)= k\, ||\mathbf{q}_1^r-\mathbf{q}_2^a||^2+ \sum_i \log^2\left( \frac{\lambda_{1,i}}{\lambda_{2,i}} \right)
\label{sim_SD_c}
\end{equation}
where the function $k$ of the anisotropies of both tensors will be exactly defined in the sequel (section \ref{sec_aniso}). This measure will be hereafter called the 'spectral quaternions' measure.\\

Table \ref{tab_cc} illustrates the numerical savings obtained by proper euclidean embedding of the desired Riemannian geometry. The saving of using quaternions as opposed to distances over rotation matrices is of the same order as the saving of using the Log-Euclidean metric instead of the affine-invariant metric. The table suggests that the numerical cost of the proposed similarity measure is competitive with the numerical cost of Log-Euclidean distance.

\begin{table}
\caption{Computational time of computing 1000 distances between a reference and random samples from a Wishart distribution. The computations are performed on a Intel Core 2 Duo 2,66 GHz with 4Go of RAM machine using a (non optimized) MATLAB code.}
\label{tab_cc}       
\begin{tabular}{llll}
\hline\noalign{\smallskip}
Affine- & Log-Euclidean & Spectral & Spectral- \\  
invariant & & & quaternions\\
\noalign{\smallskip}\hline\noalign{\smallskip}
0,47 s & 0,17 s & 0,65 s & 0,11 s\\
\noalign{\smallskip}\hline
\end{tabular}
\end{table}

\begin{figure*}[ht]
\begin{center}
\includegraphics[scale=0.75]{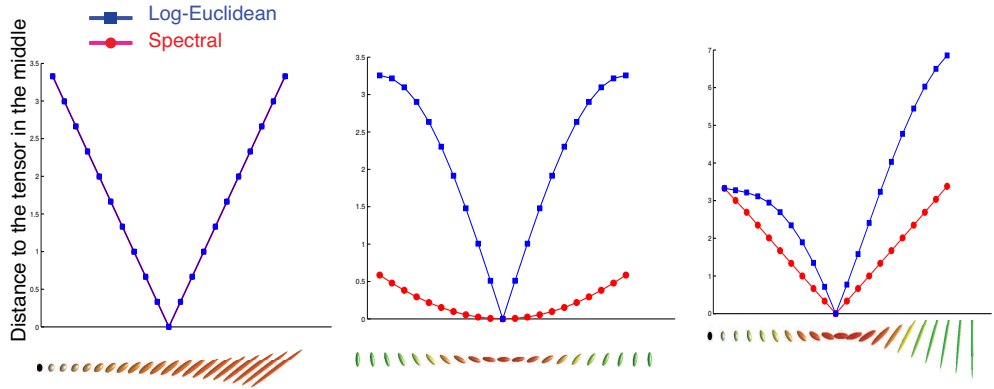}
\caption{Evolution of the different similarity measures when the properties of the tensors are smoothly varied. \emph{Left} Variation of eigenvalues. In this case, the spectral and Log-Euclidean measures give the same results. The distances vary linearly. \emph{Center} Variation of the angle of the dominant eigenvector. While the variation of the spectral distance is smooth, this is not the case for the Log-Euclidean metric. This metric appears to be more sensitive to variations in rotation, while the spectral distance is robust to these variations.  \emph{Right} Variation of both eigenvalues and angle. The Log-Euclidean measure is not regular, while the spectral quaternions distance is almost linear.}
\label{evo_measures}
\end{center}
\end{figure*}

Figure \ref{evo_measures} shows the variation of the spectral quaternions distance and the Log-Euclidean one when the characteristics of the tensors are smoothly varied. In each subfigure, the distance between each tensor of the set and the 'central' tensor of the same set is computed. When only the eigenvalues are varied (left of the figure), both the Log-Euclidean and the spectral measure give the same linear variation. In the center subfigure, only the angle of the principal eigenvector is varied. In this case, the spectral measure varies smoothly, in sharp contrast with the Log-Euclidean metric, which is very sensitive to small angle variation. The two effects are combined in the right subfigure, when both eigenvalues and orientation are varied.

\section{Anisotropy commutes with averaging}
\label{sec_aniso}
In this section, we build upon the proposed similarity measure to define an anisotropy index and an averaging operation among tensors that have the remarkable property to commute: the anisotropy of the average is the average of the anisotropies.

\subsection{Hilbert anisotropy}
Geometrically, any anisotropy scalar measure should be a scale invariant distance to identity. The Hilbert metric \citep{Koufany:2006uq} is precisely a projective distance that can be defined in arbitrary cones. It leads to the following definition that we refer to as Hilbert anisotropy (HA) in the sequel: 
\begin{equation}
\mathrm{HA}= d_{\mathrm{H}}(\mathbf{S}, \mathbf{I})= \log (\frac{\lambda_{\max}}{\lambda_{\min}})
\label{HA}
\end{equation}
where $\lambda_{\max}$ is the maximum eigenvalue of $\mathbf{S}$ and $\lambda_{\min}$, the minimum one. 
The HA index possesses all the required properties for an anisotropy index, \emph{i.e.}
\begin{itemize}
\item HA $\geq 0$ and HA $=0$ only for isotropic tensors.
\item HA is invariant to rotations: $\mathrm{HA}(\mathbf{S}) = \mathrm{HA}(\mathbf{USU}^T)$ for all $\mathbf{U} \in \mathrm{O(3)}$.
\item HA is invariant by scaling, $\mathrm{HA}(\mathbf{S}) = \mathrm{HA}(\alpha \mathbf{S})$, $\forall \alpha \in \mathbb{R}_+$ (it means that anisotropy only depends on the shape of the tensor and not on its size).
\item HA is a dimensionless number. This property is desirable and natural, as the anisotropy of the tensor physically reflects the microscopic anisotropy of the tissues, which is independent from the diffusivity.
\end{itemize}

\begin{figure}[ht]
\begin{center}
\includegraphics[scale=0.4]{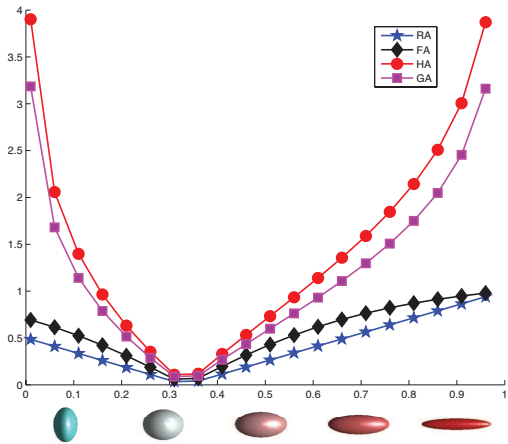}
\caption{Evolution of four indices of anisotropy with respect to modifications of eigenvalues. The difference between 'linear' indices of anisotropy (FA, $\diamondsuit$ and RA, $\bigstar$) and 'geometric' ones (HA, $\bigcirc$ and GA,$\square$) is clearly visible. The eigenvalues are given by $t, (1-t)/2, (1-t)/2,\, t \in ]0,1]$, \cite{Batchelor:2005ys}. }
\label{Ani_ind}
\end{center}
\end{figure}

Figure \ref{Ani_ind} illustrates a comparison of HA with three popular anisotropy indices: fractional anisotropy (FA), relative anisotropy (RA) \citep{Basser1996209}, the geodesic anisotropy (GA)\citep{Fletcher07riemanniangeometry}. This figure is constructed with eigenvalues equal to $t, (1-t)/2, (1-t)/2,\, t \in ]0,1]$, as explained in \citep{Batchelor:2005ys}. The diffusion tensor varies from planar to spherical for $t \in ]0, 1/3]$ and then tends to become unidirectional  when $t$ increases.\\

We use the HA index to define a weighting factor $k$ in \eqref{sim_SD_c}. The empirical choice adopted in this paper and that can be customized by the user is 
\begin{equation}
k (\mathrm{HA}_1,\mathrm{HA}_2)= \frac{1+ \tanh(3 \mathrm{HA}_1 \mathrm{HA}_2-7)}{2}.
\label{functionk}
\end{equation}
which smoothly varies from zero when one of the tensor is isotropic to one when the two tensors are strongly anisotropic. 

\subsection{Weighted mean of Diffusion Tensors}

The Riemannian mean of a set of tensors is defined as the tensor that minimizes the total squared distance to all tensors (see \citep{Fletcher07riemanniangeometry}). Here we use our similarity measure to provide a numerically tractable approximation of this quantity. 
\begin{itemize}
\item \emph{Weighted mean of two tensors}
Consider two tensors $\Su$ and $\Sd$ with respective weights $w_1\geq 0$ and $w_2 \geq 0$ ($w_1+w_2=1$).
\begin{itemize}
\item The eigenvalues of the mean tensor are defined as 
\begin{equation} \lambda_{\mu,i}= \exp( w_1 \log \lambda_{1,i} + w_2 \log \lambda_{2,i}) \label{2inter_eigv} \end{equation}
which corresponds to a Riemannian mean on $\mathrm{D}^+(3)$ for the logarithmic distance (\emph{i.e.} a geometric mean of positive numbers).
\item The mean orientation is defined by means of the mean quaternion 
\[ \mathbf{q}_m= w_1 \mathbf{q}_1 + w_2\mathbf{q}_2\]
\begin{equation} \mathbf{q}_{\mu}= \frac{\mathbf{q}_m}{|| \mathbf{q}_m||}\label{2inter_rot}\end{equation}
which is a weighted chordal mean in the space of quaternions. Note that $\mathbf{q}_1$ and $\mathbf{q}_2$ in \eqref{2inter_rot} must be realigned quaternions as in \eqref{dist_max}. 

\end{itemize}
 \item \emph{Weighted mean of more than two tensors} Consider $N$ tensors $\mathbf{S}_1, \mathbf{S}_2, \dots, \mathbf{S}_N$ ($N>2$) with respective weights $w_1, w_2, \dots w_N, \sum_i w_i =1$. Using our similarity measure, we propose a method that provides a well-defined and commutative mean. 
 \begin{itemize}
 \item Similarly to the case of two tensors, the eigenvalues of the mean are defined as the geometric mean of all eigenvalues, i.e. 
\begin{equation} \Lambda_{\mu}= \exp( \sum_{i=1}^N w_i \log(\Lambda_i) )\label{eigvN}  \end{equation}
\item To make the definition of the mean unique, we must select a quaternion, $\mathbf{q}^r$, which will serve as a reference for the realignment of all other quaternions. To this end, we first compute the weighting factors $k_i=k(\mathrm{HA}_i,\mathrm{HA}_\mu) \,, \, \forall i=1, \dots, N$. We then choose as a reference the  tensor $\mathbf{S}_r$ for which the product $w_r k_r$ is the highest (\emph{i.e.} the most informative tensor).  The realignment procedure is then applied to all the other tensors, and the realigned quaternions are denoted $\mathbf{q}_{i,r}$. 
\item The mean orientation is defined as
\[ \mathbf{q}_m= \frac{\sum_{i=1}^N w_i k_i \mathbf{q}_{i,r}}{\sum_i k_i}\]
\begin{equation} \mathbf{q}_{\mu}= \frac{\mathbf{q}_m}{|| \mathbf{q}_m||}\label{eigRN} \end{equation}
The mean is commutative, because the reference quaternion $\mathbf{q}^r$ does not depend on the labeling order of the tensors. 
\end{itemize}
\end{itemize}

The \emph{chordal} mean \eqref{2inter_rot},\eqref{eigRN} of quaternions is the solution \citep{Dai:2010fk}  of 
\begin{equation}
\mathbf{R}= \mathrm{arg}\min_{\mathbf{R}}\sum_{i=1}^N d^2_{\mathrm{quat}}(\mathbf{R}, \mathbf{R}_i) 
\end{equation}
where $d_{\mathrm{quat}}$ is the quaternion distance between rotations and $\mathbf{R}_i$ are the rotations to be averaged. It is thus a Riemannian mean over the set of rotations for the chordal distance defined on the set of quaternions. Since the quaternion distance is a good approximation of the geodesic distance between rotations for small rotations, the quaternion mean is viewed as a good approximation of the real mean rotation.

\subsection{Properties of the mean}
The proposed spectral-quaternion mean (denoted $\mathbf{S}_{\mu}$ in the following) enjoys two `information- preserving' properties of interest for DTI processing.

\begin{itemize}
\item \emph{Determinant of $\Sm$} From Eq. \eqref{2inter_eigv}, one easily shows that
\[ \det(\mathbf{S}_{\mu})= \exp(w_1 \log(\det(\Su))+ w_2 \log(\det(\Sd)))\] \emph{i.e.} the determinant is the geometric mean of the determinants of the tensors. The same property holds in the case of $N>2$ tensors, thanks to Eq. \eqref{eigvN}.
\[ \det(\mathbf{S}_{\mu}) = \exp(\sum_{i=1}^N w_i \log(\det(\mathbf{S}_i)))\] 
The Log-Euclidean mean shares the same property \citep{arsigny:328}. This ensures that there is no swelling effect (increase of the volume of the ellipsoid) during the averaging of tensors.

\item \emph{Anisotropy of $\Sm$}
The anisotropy of the mean is the arithmetic mean of the anisotropies of the tensors. In other words, anisotropy commutes with averaging of tensors. Indeed, if HA$_1$ is the anisotropy of $\Su$ and HA$_2$ the one of $\Sd$, we have that
\begin{eqnarray*}
\mathrm{HA}_{\mu} &=& \log\left(\frac{\lambda_{\mu,1}}{\lambda_{\mu,3}}\right) \\
     &=& w_1 \log\left(\frac{\lambda_{1,1}}{\lambda_{1,3}}\right)+ w_2 \log\left(\frac{\lambda_{2,1}}{\lambda_{2,3}}\right)\\
     &=& w_1 \mathrm{HA}_1+ w_2 \mathrm{HA}_2 
\end{eqnarray*}
In the case of more than two tensors, it can easily be shown that 

\[ \mathrm{HA}_{\mu} = \sum_{i=1}^N w_i \mathrm{HA}_i \] 
This important property is not satisfied by the Log-Euclidean method, neither by the affine-invariant mean. It is because of this property that the proposed metric addresses the shortcomings of the affine invariant metric pointed in Figure \ref{fig1}.

\end{itemize}

\section{Information preserving interpolation}
As a direct application of the weighted means introduced in the preceding section, we propose an anisotropy preserving interpolation scheme for diffusion tensor images. Other examples of application of weighted means include the computation of statistics about group of tensors.\\

As already mentioned in \citep{Zhang2006764,kindlmannMICCAI07,arsigny:328}, an adequate interpolation method is important for the processing of diffusion tensor images and particularly for the extension of usual registration techniques (for scalar images) to the case of tensor images. This interpolation scheme is necessary to resample images. Here, we provide a direct generalization of classical interpolation method, where the interpolated value is computed as the weighted mean of the original tensors. Equations \eqref{2inter_eigv} and \eqref{2inter_rot} directly provide interpolating curves in the space of diffusion tensors, with, if $t$ is the parameter of the interpolation, $w_1= (1-t)$ and $w_2= t$. Those curves are reasonable approximation of geodesic curves employed in Riemannian interpolation theory \citep{Fletcher07riemanniangeometry}.  \\

\begin{figure*}[ht]
\begin{center}
\includegraphics[scale=0.7]{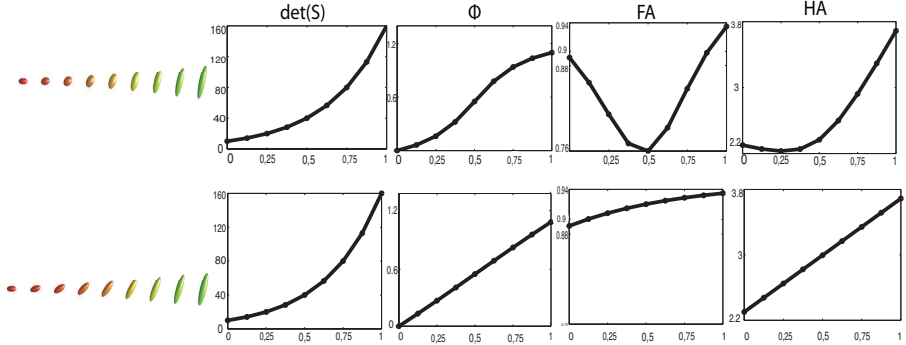}
\caption{Geodesic interpolation between two tensors. \emph{Top}: Log-Euclidean interpolation. \emph{Bottom}: Spectral-quaternion interpolation. Clear differences can be seen between these two methods, namely for the evolutions of degrees of anisotropy.}
\label{interp2t}
\end{center}
\end{figure*}

The scheme for the bi and tri-linear interpolations of scalars can not been extended to tensors, because unlike linear interpolations, interpolation along curves does not commute. This is why the commonly used solution is to compute interpolation through a weighted average of diffusion tensors \citep{citeulike:451696,Fletcher07riemanniangeometry,Arsigny:2006fk}. The weight associated to each tensor is a function of the grid distance between this tensor and the location of the interpolated tensor. In this work, if $(x_1,x_2,x_3) \in [0,1]\times [0,1] \times [0,1]$ are the coordinates of the interpolated tensor and $(\alpha_1, \alpha_2, \alpha_3) \in \{0,1\}\times \{0,1\} \times \{0,1\} $ the coordinates of the point $\alpha$ of the grid, the following function will be used
\[ w_{\alpha}(x_1,x_2,x_3)= \prod_{i=1}^3 (1-\alpha_i + (-1)^{1-\alpha_i} x_i).\]
Figure \ref{interp2t} shows the curve interpolation between two tensors using both the Log-Euclidean and the spectral quaternions frameworks. As in \citep{Zhou:2010fk}, the variation of the main information conveyed by the tensors is also shown. As previously shown, the Hilbert anisotropy is linearly interpolated by the novel framework, while this information is significantly degraded in the Log-Euclidean method. A similar behavior is found for the evolution of the fractional anisotropy. Both methods geometrically interpolate the determinant. It is also interesting to analyse the difference in $\phi$, the angle between the first eigenvector of the first tensor and the first eigenvector of the weighted mean. While the spectral measure produces a quasi linear interpolation of this angle, this is not the case for the Log-Euclidean framework.

\begin{figure*}[ht]
\begin{center}
\includegraphics[scale=0.7]{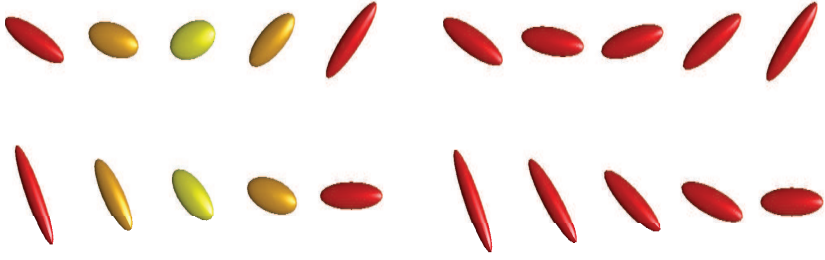}
\caption{Geodesic interpolation between two tensors, illustration of the invariance by rotation. \emph{Left}: Log-Euclidean interpolation. \emph{Right}: Spectral quaternions interpolation. The colors of tensors are based on the anisotropy. From top to bottom, the original tensors are rotated. With both frameworks, it is clear that the interpolated tensors are rotated in the same way, which means these frameworks are invariant by rotation. However, a clear difference can be seen between the frameworks, since the spectral quaternions one preserves the anisotropy, while the Log-Euclidean one produces a degradation of this information.}
\label{interp2trot}
\end{center}
\end{figure*}

Figure \ref{interp2trot} illustrates the invariance by rotation of the curve interpolation of the spectral quaternions framework. Although the interpolations are different (in particular for the shape of the tensors), the Log-Euclidean framework also possesses this property. 

\begin{figure*}[ht]
\begin{center}
\includegraphics[scale=0.6]{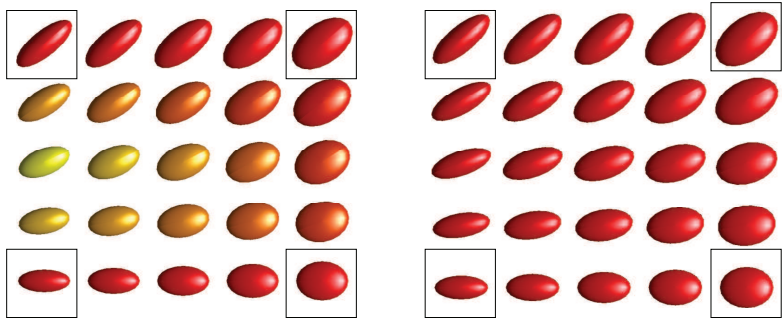}
\caption{Geodesic interpolation between four tensors at the corners of the grids. \emph{Left}: Log-Euclidean interpolation. \emph{Right}: Spectral interpolation. Colors of the ellipsoids indicates their anisotropy from red (high anisotropy) to yellow for smaller anisotropies. While the anisotropy is equal for each tensor computed by the spectral quaternions method, this is not the case with the Log-Euclidean one, where a decrease of the anisotropy can be seen.}
\label{interp4t}
\end{center}
\end{figure*}

Using the method described above for computing the weighted means of many tensors, the interpolation of four tensors at the corners of a grid can be computed, as illustrated in Figure \ref{interp4t}, where colors of the tensors is determined by HA. The four tensors at the corners have equal Hilbert Anisotropy. As a consequence, all tensors interpolated by the spectral method conserve the same anisotropy, while this is not the case with the Log-Euclidean method. 

\section{Conclusions}
In this paper, we have introduced a novel geometric framework for the processing of Diffusion Tensor Images. This framework is based on the spectral decomposition of tensors. The main advantage of this method is to preserve the anisotropy of tensors during the processing. Moreover, it possesses all the important properties of existing metrics, such as the invariances and the preservation of other information as the determinant and the orientation of tensors. The preservation of anisotropy is rooted in the spectral decomposition of the tensors, which allows for comparing separately rotations and the spectral information.\\
Computational obstacles previously faced in similar attempts \citep{10.1109/CVPR.2001.990631,chefdhotel04} are circumvented by embedding the set of rotation matrices in the space of quaternions, long used for its numerical efficiency in robotics.\\
The resulting interpolation method retains the computational tractability and the geometry of the Log-Euclidean framework but addresses a limitation of this framework regarding the degradation of anisotropy.\\
Although several illustrations of the paper exemplify the potential benefit of preserving anisotropy through averaging and interpolation operation encountered in statistical process, the benefits of the proposed framework remain to be demonstrated on real data. Registration and tractography are two particular areas where the advantages of the proposed method should be evaluated quantitatively.

\bibliographystyle{spbasicemph}      
\bibliography{biblio_AP_paper}   

\end{document}